\documentclass[nohyperref, twocolumn]{article}

\usepackage{natbib}
\usepackage{microtype}
\usepackage{graphicx}
\usepackage{subfigure}
\usepackage{booktabs} % for professional tables
\usepackage{algorithmic}
\usepackage{algorithm}
\usepackage{bm, bbm, amsmath}
\usepackage{hyperref}

\usepackage{amsmath}
\usepackage{amssymb}
\usepackage{mathtools}
\usepackage{amsthm}

\usepackage[capitalize,noabbrev]{cleveref}

\theoremstyle{plain}

\theoremstyle{definition}

\theoremstyle{remark}

\title{Scalable Bayesian Optimization Using Vecchia Approximations of Gaussian Processes}

\date{%
    $^1$Department of Statistics, Texas A\&M University\\%
}
\author{Felix Jimenez$^1$\thanks{\texttt{felix.jimenez@tamu.edu}} \and Matthias Katzfuss$^1$ \thanks{Corresponding author:\texttt{katzfuss@gmail.com}}}

\begin{document}

\maketitle
\begin{abstract}
Bayesian optimization is a technique for optimizing black-box target functions. At the core of Bayesian optimization is a surrogate model that predicts the output of the target function at previously unseen inputs to facilitate the selection of promising input values. Gaussian processes (GPs) are commonly used as surrogate models but are known to scale poorly with the number of observations. We adapt the Vecchia approximation, a popular GP approximation from spatial statistics, to enable scalable high-dimensional Bayesian optimization. We develop several improvements and extensions, including training warped GPs using mini-batch gradient descent, approximate neighbor search, and selecting multiple input values in parallel. We focus on the use of our warped Vecchia GP in trust-region Bayesian optimization via Thompson sampling. On several test functions and on two reinforcement-learning problems, our methods compared favorably to the state of the art.
\end{abstract}


\section{Introduction \label{sec:intro}}
Bayesian optimization (BO) is a class of techniques for black-box function optimization \citep{Mockus1989}. BO has found success in a variety of areas, including reinforcement learning \citep[e.g.,][]{Calandra2016}, tuning of machine-learning algorithms \citep[e.g.,][]{Swersky2013}, A/B testing \citep[e.g.,][]{Letham2019} and material discovery \citep[e.g.,][]{gomez2018automatic}.

The basic idea of BO is to approximate the black-box function with a cheap-to-evaluate surrogate, typically a GP, and perform optimization with the surrogate. Starting with an initial collection of function evaluations, the trained GP is combined with a heuristic to generate an \textit{acquisition function}, which assigns values to unseen inputs. The input with the highest acquisition value is evaluated and the result is added to the dataset. This process continues iteratively until a predefined stopping criterion is met or the computational budget is depleted. For a more complete review of Bayesian optimization see \citet{Frazier2018AOptimization}.

As most BO techniques use a GP surrogate, they inherit the scalability issues of GPs. It becomes expensive to search a high-dimensional space when optimizing the acquisition function.  

This work addresses this scalability issue by extending the Vecchia GP approximation from spatial statistics \citep[e.g.,][]{Vecchia1988,Katzfuss2017a} to work in high dimensions with many observations. Through these extensions, we demonstrate how Vecchia GPs can be used in concert with existing BO procedures. To be precise, our contributions are: 
\begin{itemize}
    \item Introduce Vecchia GP approximations to the machine-learning community.
    \item Adapt Vecchia GPs to BO and improve their flexibility and scaling with warped kernels, vector quantization, approximate nearest neighbors, minibatch training and variance correction. 
    \item Empirically demonstrate the utility of Vecchia GPs within existing BO frameworks. 
\end{itemize}

\section{Existing Solutions}\label{sec:related}
There has been interest in scaling BO in two ways: higher input dimensions and more function evaluations. While we mostly focus on the latter here, these two aspects are related, in that successful high-dimensional optimization typically requires many function evaluations. 

High-dimensional BO is thoroughly studied with many proposed solutions, including transformations to a lower-dimensional space \citep{gomez2018automatic}, assuming an additive form of the function \citep{kandasamy2015high}, and using sparsity-inducing priors \citep{eriksson2021high}.

Scaling BO to many function evaluations is less explored. Existing approaches rely on additivity assumptions \citep[e.g.,][]{wang2018batched} or GP approximations \citep{mcintire2016sparse, maddox2021conditioning}. The Trust Region BO method \citep[TuRBO;][]{eriksson2019scalable} has also seen great success by relying on Thompson sampling and limiting the input search to trust regions, which are determined by the lengthscale parameters of the GP.

Here we argue and show that GP-based BO methods can be made scalable via the Vecchia approximation, and we establish that existing scalable BO methods can be improved by using Vecchia GPs instead of inducing point methods \citep[e.g.,][]{Hensman2013GaussianData}, which can suffer from important limitations \citep{Stein2013a}. We hope to provide readers with a set of techniques for incorporating Vecchia GPs into existing BayesOpt procedures to improve their scalability.

\section{Background}\label{sec:background}

\subsection{Review of Bayesian Optimization \label{sec:bayesopt}}

Bayesian optimization aims to solve the problem
\begin{equation}
\label{eq:opt}
\displaystyle \min_{\bm x \in \mathcal{X}} f(x),    
\end{equation}
for a ``black-box'' objective function $f$ (called the target function) over some input space $\mathcal{X}\subset \mathbb{R}^d$. In BayesOpt, $f$ is often modeled as a Gaussian process (GP), $f(\cdot) \sim GP(\mu_{\bm \theta},K_{\bm \theta})$, whose mean $\mu_{\bm \theta}$ and kernel function $K_{\bm \theta}$ depend on hyperparameters $\bm \theta$.

Function evaluations $y_i = f(\bm x_i)$ are obtained sequentially. Given the first $n$ observations $\bm y_{1:n}$ at inputs $\bm x_{1:n}$, the goal is to find a new input $\bm x_{n+1}$, such that the global optimum of $f$ is likely at $\bm x_{n+1}$. 

This is usually achieved by maximizing an acquisition function, $\bm x_{n+1} = \arg\max_{\bm x \in \mathcal{X}} a_{\bm \theta^*}(\bm x| \bm y_{1:n})$, which is a function that assigns a measure of value to unseen locations based on a utility function, and where $\bm \theta^* = \arg\max_{\bm \theta} \log p_\theta(\bm y_{1:n})$ is the maximum likelihood estimate.

Alternatively, new inputs can be chosen using  Thompson sampling \citep{Thompson1933}, which has been shown to work well empirically \citep{Chapelle2011}. In Thompson sampling, we consider a collection of $M$ inputs, $\bm X = \bm x_1^*,\ldots,\bm x_M^*$, sampled from the input space $\mathcal{X}$. We generate a joint sample from the posterior at these locations $(y_1^*,\ldots,y_M^*) \sim p_{\bm \theta}(f(\bm X)|\bm y_{1:n})$. We choose $\bm x_{n+1}$ to be the input value with the lowest sampled value, $\bm x_{n+1} = \bm x_{j}^\star$ with $j = \arg\min_{i\in \{1, ..., M\}} y_i^*$. If we want a batch of $q>1$ query points, we generate $q$ independent batches from the posterior over the same set $\bm X$ and choose the lowest value from each batch (avoiding duplicates).

\subsection{Vecchia GP Approximations\label{sec:vecchiareview}}

Given $n$ GP observations $\bm y_{1:n}$ as described above, the Vecchia GP approximation \citep{Vecchia1988, Katzfuss2017a} uses a modified likelihood to turn the $\mathcal{O}(n^3)$ complexity for standard GP regression into $\mathcal{O}(n m^3)$, with $m \ll n$. (A further reduction to $\mathcal{O}(n m^2)$ is possible via more complicated grouping methods described in \citealp{Schafer2020}.) The following provides details on Vecchia GPs, including the likelihood and the posterior predictive distribution. 

\subsubsection{Vecchia Likelihood}
Recall that any joint density of $n$ observations $\bm y_{1:n}$ can be decomposed exactly as a product of conditional densities: $p(\bm y_{1:n}) = \prod_{i=1}^n p(y_i|\bm y_{1:i-1})$, where $\bm y_{1:0} = \emptyset$. This motivates the Vecchia approximation \citep{Vecchia1988}: 
\begin{equation}
    \label{eq:vecchia}
\textstyle \widehat{p}(\bm y_{1:n}) = \prod_{i=1}^n p(y_i|\bm y_{c(i)}),
\end{equation}
where each $c(i) \subset \{1,\ldots,i-1\}$ is a conditioning index set of size at most $m$, and the $n$ conditional distributions $p(y_i|\bm y_{c(i)})$ are all Gaussian and can be computed in parallel, each in $\mathcal{O}(m^3)$ time. 

\subsubsection{Vecchia Properties}
Vecchia approximations have several attractive properties \citep[e.g., as reviewed by][]{Katzfuss2020}. The joint distribution $\widehat{p}(\bm y_{1:n}) = \mathcal{N}(\bm \mu,\widehat{\bm K})$ implied by \eqref{eq:vecchia} is multivariate Gaussian, where the inverse Cholesky factor $\widehat{\bm K}^{-1/2}$ is sparse with $\mathcal{O}(nm)$ nonzero entries \citep[e.g.,][]{Katzfuss2017a}. 
Under the sparsity pattern implied by the $c(i)$, the Vecchia approximation results in the optimal $\widehat{\bm K}^{-1/2}$ as measured by Kullback-Leibler (KL) divergence, $KL(p(\bm y)\| \widehat{p}(\bm y))$ \citep{Schafer2020}. 
Growing the conditioning sets $c(i)$ decreases the KL divergence \citep{Guinness2016a}; for $m=n-1$, the approximation becomes exact, $\widehat p(\bm y) = p(\bm y)$.
Thus, $m$ trades off low computational cost (small $m$) and high accuracy (large $m$); crucially, high accuracy can be achieved even for $m \ll n$ in many settings.

\subsubsection{Ordering and Conditioning Sets}

Aside from the choice of $m$, the accuracy of a Vecchia approximation depends on the ordering of $\bm y_{1:n}$ and on the choice of the conditioning sets $c(i)$ in \eqref{eq:vecchia}. 

\begin{figure*}[ht]
    \centering
    \includegraphics[width =1\linewidth]{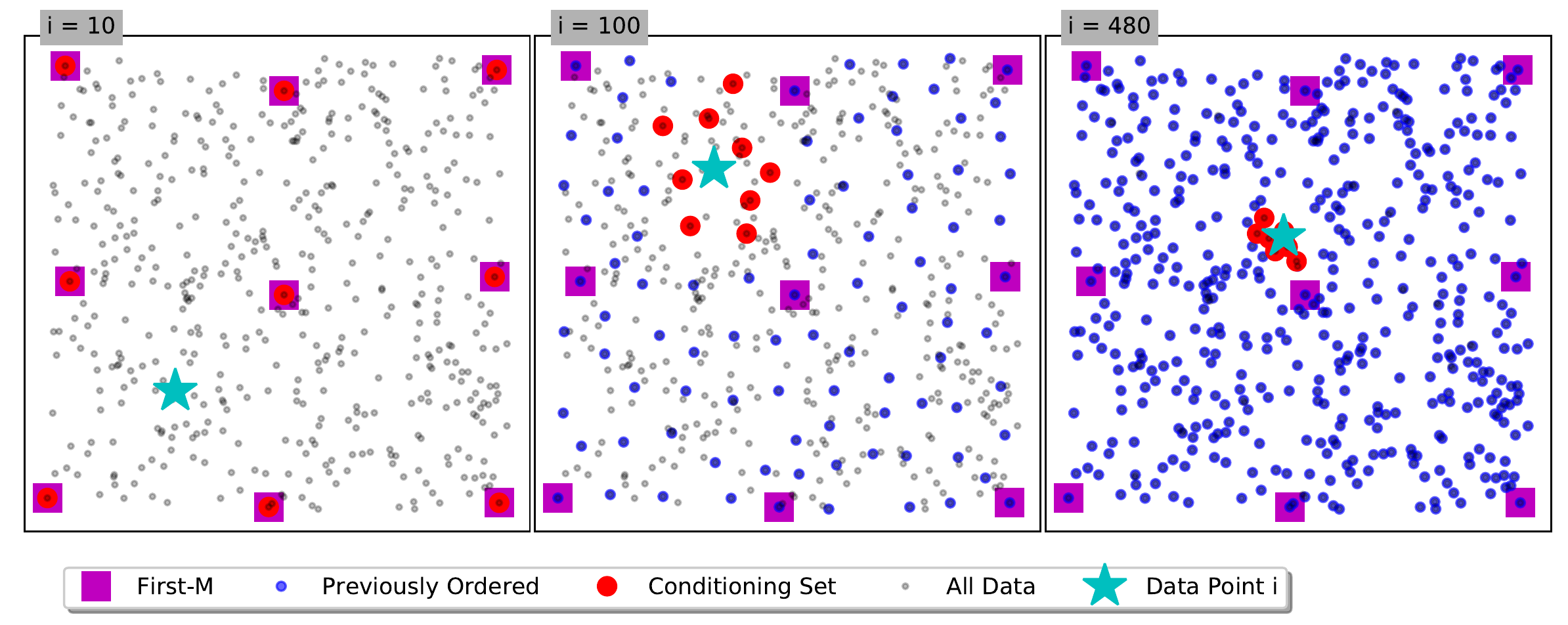}
    \caption{Illustration of maximin ordering and conditioning sets for $n=500$ uniformly sampled inputs (grey dots) on $[0,1]^2$. For $i=10$, $i=100$, and $i=480$ in the left, middle, and right panel, we show the $i$th input (cyan star) in the maximin ordering, the previously ordered $i-1$ inputs (blue dots), including the (approximate) nearest $m=9$ inputs (red circles). For increasing $i$, the first $i$ inputs form increasingly dense grids, so that the distance to the $m$ NN decreases gradually and systematically. In contrast, for fully independent conditional approximations (FIC), each input has the same conditioning set, here illustrated by the first $m$ inputs in the maximin ordering (magenta squares). Thus FIC effectively ignores many closer previously-ordered inputs, which can lead to large approximation errors.}
    \label{fig:max_min}
\end{figure*}

In our experience, the highest accuracy for a given $m$ can be achieved by combining maximum-minimum-distance (maximin) ordering with conditioning on the $m$ (previously ordered) nearest neighbors (NNs), which is illustrated in Figure \ref{fig:max_min}. Maximin sequentially picks each variable in the ordering as the one that maximizes the minimum distance to previously ordered variables. Exact maximin ordering and NN conditioning can be computed in quasilinear time in $n$, and $m$ needs to grow only polylogarithmically with $n$ for $\epsilon$-accurate Vecchia approximations with maximin ordering and NN conditioning under certain regularity conditions \citep{Schafer2020}.

In Figure \ref{fig:max_min}, we see that for index $i$ in the maximin ordering, the $m$ previously ordered NNs can be far away for small $i$ and very close for large $i$. This global and local behavior helps Vecchia both learn accurate values for hyperparameters and make good predictions. This is in contrast to low-rank approximations that effectively use the same conditioning sets for each $i$ and hence focus on global behavior at the cost of ignoring local structure.
\subsubsection{Vecchia Predictions\label{sec:pred}}
The Vecchia approximation can also be used to obtain an accurate approximation of the joint posterior predictive distribution of an unobserved $n_p$-vector $\bm y_p$ \citep{Katzfuss2018} of the form 
\begin{equation}
\label{eq:vecchiapred}
\textstyle \widehat p(\bm y_p|\bm y_{1:n}) = \prod_{i=1}^{n_p} p(y_p^{(i)}|\bm y_{c_p(i)}) = \mathcal{N}(\bm\mu_p,(\bm L^T\bm L)^{-1}),
\end{equation}
where $\bm y_{c_p(i)}$ denotes the $m$ NNs of $y_p^{(i)}$ among $\bm y_{1:n}$ (or $\bm y_{1:n} \cup \bm y_p^{(1:i-1)}$, if joint predictions are desired). The distribution in \eqref{eq:vecchiapred} can be computed in $\mathcal{O}(n_pm^3)$ time, for example using the univariate Gaussian conditional distributions in the middle part of \eqref{eq:vecchiapred}.

Alternatively, the matrix expression for the posterior predictive distribution on the right-hand side of \eqref{eq:vecchiapred} can be computed by first filling in the elements of a sparse matrix $\bm U$, as follows. Let $g(i)$ be the indices of the conditioning set for observation $i$ with input $\bm x_i$ and response $y_i$. The $(j,i)$th element of $\bm U$ is given by:
\begin{equation}
\bm{U}_{j,i}=
    \begin{cases}
        d_i^{-1/2}, & i=j\\
        -b_i^{(j)}d_i^{-1/2}, & j\in g(i)\\
        0, & \text{otherwise}, 
    \end{cases}
\end{equation}
where $\textbf{b}_i^T = K(\bm x_i, \bm x_{g(i)})K(\bm x_{g(i)}, \bm x_{g(i)})^{-1}$, $d_i = K(\bm x_i, \bm x_i) - \bm b_i^T K(\bm x_{g(i)}, \bm x_i)$. We don't need to construct the entire matrix $\bm U$, but instead we simply need to fill in sub parts of the matrix for posterior inference. In particular let $\bm U_{p,p}$ denote the sub-matrix of $\bm U$ corresponding to prediction locations and $\bm U_{n,p}$ denote the sub-matrix for training locations and prediction locations. We then use these sub-matrices to compute the posterior mean, 
\begin{align*}
    \bm \mu_p = (\bm U_{p,p}^T)^{-1}(\bm U_{n,p})^T \bm y_{1:n},
\end{align*}
and the Cholesky factor of the posterior precision is simply given by $\bm L = \bm U_{p,p}^T$.
The matrices $\bm U_{p,p}$ and $\bm U_{n,p}$ are both very sparse with at most $m$ non-zero elements per column, reducing the cost of the operations. We provide additional details on Vecchia predictions in Appendix \ref{appendix:UMatrix}. 

\subsection{Vecchia and SVI-GP}\label{sec:approximationCompare}

\citet{Hensman2013GaussianData} showed how inducing point GP approximations can be trained using stochastic variational inference, and we refer to this model as SVI-GP. To demonstrate how Vecchia GPs perform on functions with steep regions, we compared Vecchia GP to SVI-GP on the Michalewicz function \citep{molga2005test}. 

We created training data by generating $n=500$ observations from the 15-dimensional version of the function and adding independent noise from $\mathcal{N}(0, 0.05 ^ 2)$. We then generated a test set mimicking a typical BO prediction setting by finding the minimum in the training set and generating 50 noiseless realizations in a neighborhood of the minimum. 

For a range of values for $m$, we fitted an exact Gaussian process, the Vecchia approximation and the SVI-GP, and computed all the posteriors on the test set. Then we computed the KL-divergence $KL(p(\bm y_p|\bm y_{1:n})\| \widehat{p}(\bm y_p|\bm y_{1:n}))$ between the exact posterior and each of the approximate posteriors. We repeated this process, including generating new data, and averaged the results as we varied $m$. For Vecchia, $m$ corresponds to the number of NN, and for SVI-GP, $m$ corresponds to the number of inducing points.

\begin{figure}[ht]
    \centering
    \includegraphics[width =.9\linewidth, trim = {0 1.0cm 0 0.4cm}]{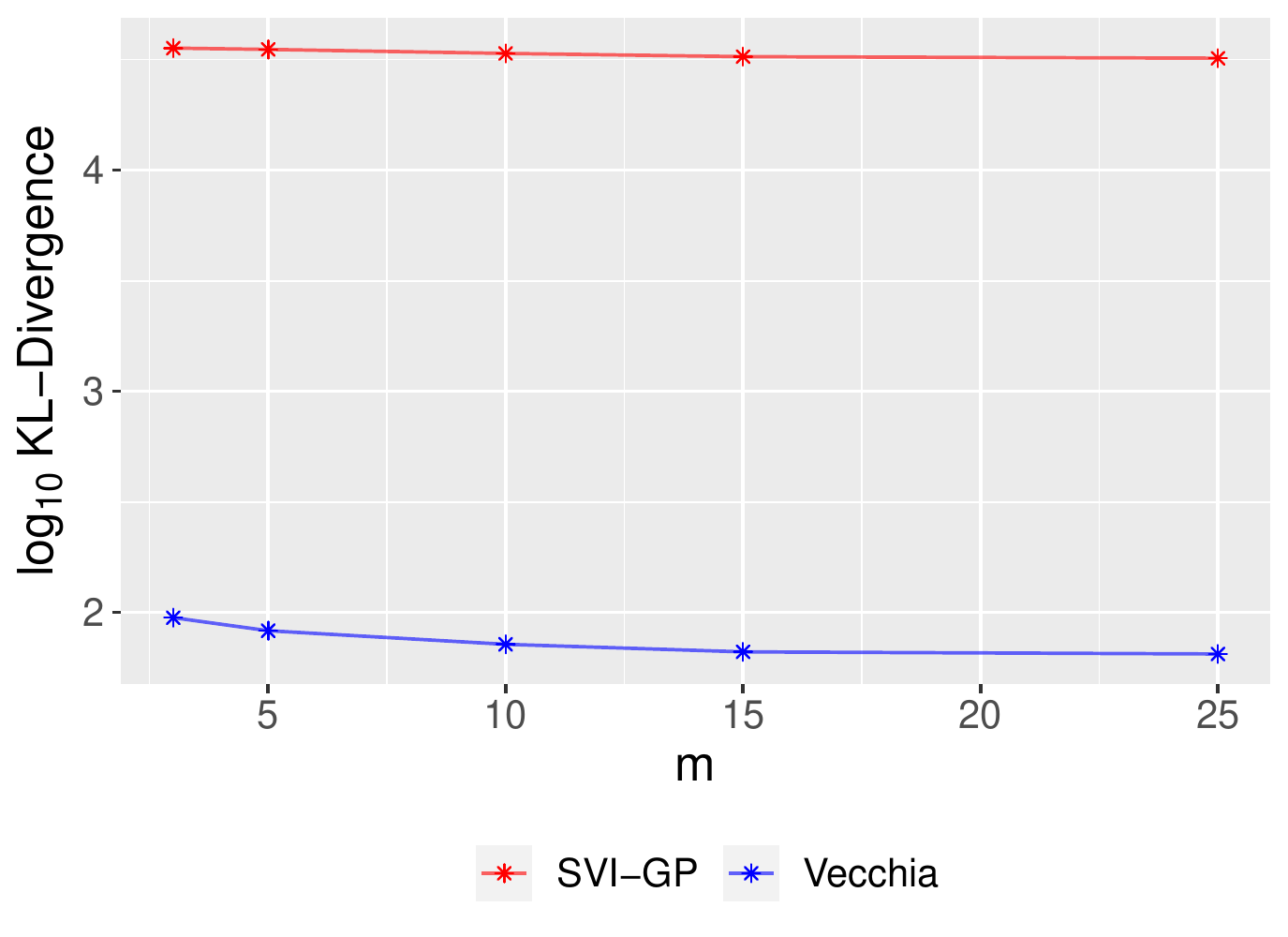}
    \caption{For prediction of the 15-dimensional Michalewicz function, the log of the KL-divergence between an exact GP and approximate GPs based on the Vecchia approximation (blue) and the SVI-GP approximation (red). }
    \label{fig:kl_vecchia_SVIGP}
\end{figure}

Figure \ref{fig:kl_vecchia_SVIGP} shows the log of the KL-divergence for both models as we increase $m$. In this example, Vecchia outperformed SVI-GP for all values of $m$ by several orders of magnitude. Based on further exploration, we conjecture that Vecchia more closely matches the local behavior of the function at the optima and SVI-GP is more prone to return estimates similar to the average over the entire domain. This success at modeling local fluctuations makes Vecchia a strong candidate for BO, where we often have sharp spikes at the global optima. 

\section{Vecchia GPs for Bayesian optimization\label{sec:vecchBayesOpt}}
In this section we detail our modifications to Vecchia GPs to improve their performance for BO.

\subsection{Warped Kernel}

A flexible GP kernel can be obtained by taking a base kernel $\tilde K$ (e.g., Mat\'ern 5/2) and warping the inputs with a nonlinear function $w(\cdot)$,
\begin{equation}
    K(\bm x,\bm x') = \tilde K(\|w(\bm x)-w(\bm x')\|).
    \label{eq:warpingkernel}
\end{equation}
In particular, we warp using the Kumarswamy CDF \cite{balandat2020botorch}, which reduces to the identity for a certain choice of hyperparameter values.
To fully utilize this flexibility within Vecchia GPs, we carry out the ordering and conditioning-set selection based on Euclidean distance between the warped inputs $w(\bm x_1),\ldots,w(\bm x_n)$ instead of the original inputs $\bm x_1,\ldots,\bm x_n$.
When the training data are observed with noise, we assume $\tilde K$ includes a nugget term $\mathbbm{1}_{i = j} \tau$.

\subsection{Training Using Stochastic Gradient Descent}

The kernel $K=K_{\bm \theta}$ depends on unknown hyperparameters in the base kernel $\tilde K$ and in the warping function $w$. To learn the potentially high-dimensional hyperparameter vector $\bm \theta$, we optimize the Vecchia loglikelihood $\log \widehat{p}_{\bm \theta}(\bm y_{1:n})$ given by \eqref{eq:vecchia} using gradient descent. Importantly, on the log scale, this density is a sum of $n$ terms of the form $\log p(y_i|y_{c(i)})$, each of which can be computed in $\mathcal{O}(m^3)$ time (given the $m$ NNs). We use this to approximate the gradient of \eqref{eq:vecchia} using mini-batches of the data:
\begin{equation} \textstyle
\nabla_{\bm \theta}\widehat p(\bm y_{1:n}) \approx \sum_{j \in b} (n/|b|) \, \nabla_{\bm \theta} \log p(y_j|\bm y_{c(j)}),
\end{equation}
where $b \subset \{1,...,n\}$. 

With this change, the training time depends on the mini-batch size $b$ instead of the total data size $n$ given the ordering and NN. This reduction in time complexity is critical, as we are estimating the GP hyperparameters after each step in the BO procedure.  

\begin{figure}[ht]
    \centering
    \includegraphics[width =.9\linewidth]{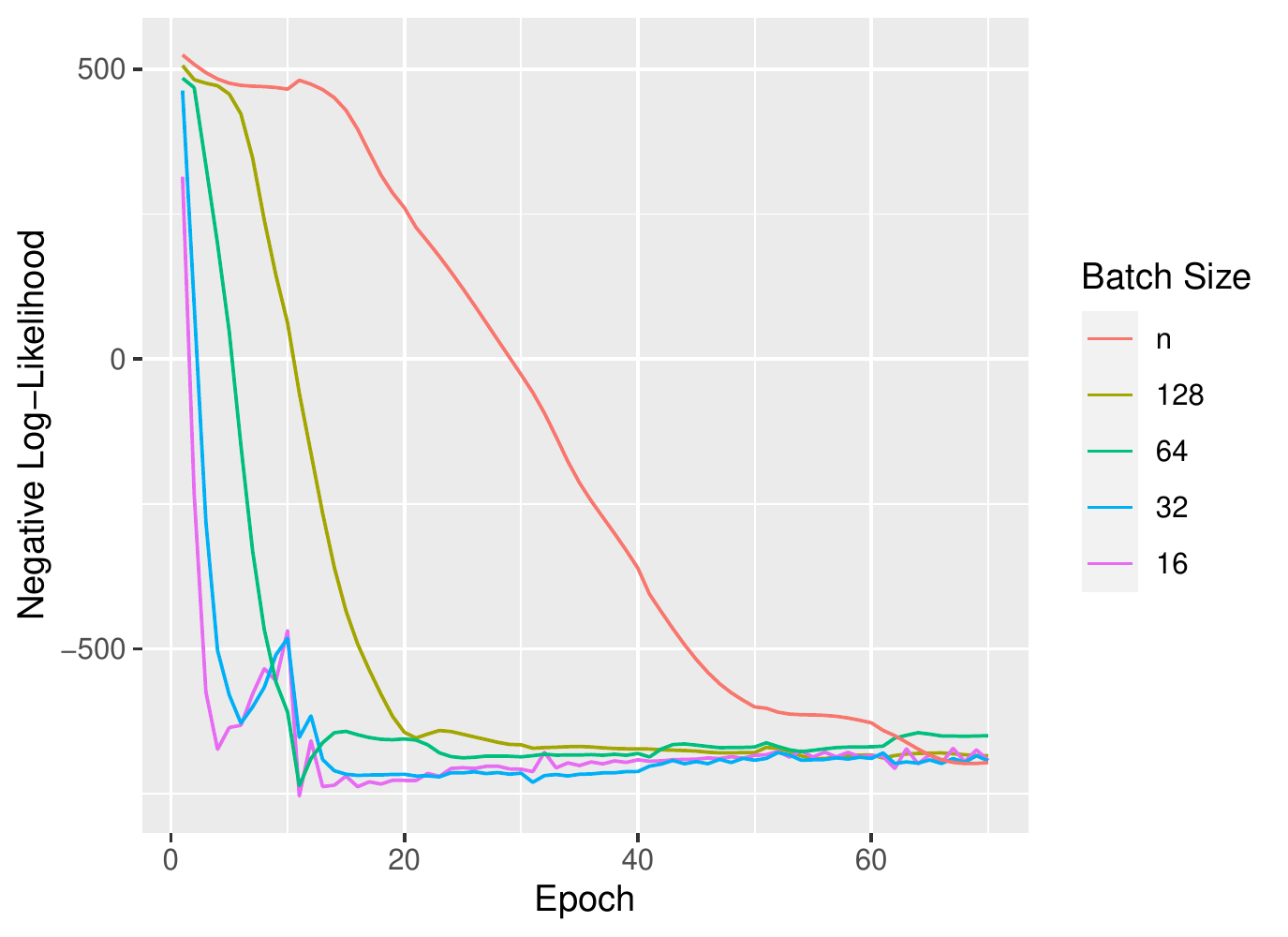}
    \caption{Log-likelihood vs. epoch for different batch sizes.}
    \label{fig:llCompare}
\end{figure}

Figure \ref{fig:llCompare} shows the negative log-likelihood vs the epoch (i.e., the number of passes through the data) during training for different batch sizes, for data generated using a simple sinusoid. This illustrates the typical training dynamics for Vecchia GPs trained using mini-batch gradient descent. For of our applications we find that a batch size of 64 offers a good balance between stable training and speed. 

\subsection{Nearest Neighbors and Ordering}
The nearest-neighbor search can become a bottleneck in the Vecchia procedure, but we have found that using approximate NN yields good performance. In particular, we suggest using approximate NNs with product quantization \citep{jegou2010product, JDH17}. This allows for balancing accuracy and speed as we consider how much data is available. As illustrated in Figure \ref{fig:max_min}, in our experience the approximate NN search is often good but not perfect for small to medium maximin index $i$ (middle panel) but it is typically highly accurate for large $i$.

For ordering data, we have found that an approximate maximin ordering works well. The data is shuffled, split into disjoint sets, ordered within each set, and recombined. Details of this procedure are given in Appendix \ref{alg:approxMaxMin}. For really large datasets, maximin ordering can even be replaced by a completely random ordering at the cost of a slight decrease in accuracy for a given $m$ \citep{Guinness2016a}.

Together, these changes result in Vecchia approximations that can be applied to 1,000-dimensional GPs with $n=100{,}000$ observations and beyond.

\subsection{Calibration of the Prediction Variance}
To improve uncertainty quantification for out-of-sample predictions, we calibrate the marginal predictive variances as in \citet{Katzfuss2020}. In short, we maximize the log-probability of a hold-out set by adding a term to the diagonal of the covariance matrix. 

To do this we first obtain an estimate $\bm \theta^*$ of the hyperparameters based on all available data. We then create a hold-out set, $\tilde{\bm y}$, and compute the posterior mean $\tilde{\bm \mu}_{\bm \theta^*}$ and variance $\tilde{\bm \Sigma}_{\bm \theta^*}$ for this hold-out set based on the training data not in the hold-out set, as described in Section \ref{sec:pred}. Finally, we choose the variance inflation factor, $b_v$, to maximize the posterior log-probability of this holdout set:
\begin{equation}
\displaystyle \max_{b_v \in [0, 2]} \text{ log } p(\tilde{\bm y}|\tilde{\bm \mu}_{\bm \theta^*}, \tilde{\bm \Sigma}_{\bm \theta^*} + b_v \bm I_{\tilde n}).
\end{equation}
The details of this procedure are given by Algorithm \ref{alg:vecchVariance}, where the funtion $KNN(\bm a,\bm b,c)$ returns the $c$ NN of $\bm b$ in $\bm c$ and the function $subsample(\bm i,  j)$ selects a subsample of size $j$ from $\bm i$. 

With $b_v>0$, the predictive variance is inflated, which encourages more exploration in the resulting BO procedure. Since the observations are processed to have zero mean and variance equal to one, the value of $b_v$ is generally at most 1.4 but usually less than 0.1. In our experience, this variance correction is helpful when the observations are noisy or the target function is highly variable. However, if the function is smooth and we have noise-free observations, simply using the initial Vecchia predictions is sufficient.

\begin{algorithm}
\caption{Vecchia Variance Calibration}
\label{alg:vecchVariance}
\begin{algorithmic}
   \STATE {\bfseries Input:} data $\textbf{x}, \textbf{y}$, parameters $\bm \theta$
   \STATE $I_{x^*} \gets \arg \max(\textbf{y})$
\STATE $I_{O} \gets KNN(\textbf{x}[I_{x^*}], \textbf{x}, 5 * q)$
\STATE $I_{O} = subsample(I_{O}, q)$
\STATE $\bm \tilde{X} = \textbf{x}[I_O] \cup \textbf{x}[I_{\bm x^*}]$
\STATE $\bm \tilde{y} = \textbf{x}[I_O] \cup \textbf{y}[I_{\bm x^*}]$
\STATE $\tilde{\mathcal{D}} = \{\textbf{x} \setminus \bm \tilde{X}, \textbf{y} \setminus \bm \tilde{y}\}$
\STATE $\mathcal{L}_{b_v} = p(\bm \tilde{y} ; \tilde{\bm \mu}_{\bm \theta^*}, \tilde{\bm \Sigma}_{\bm \theta^*})$

\STATE {\bfseries Return:} $\arg\max_{b_v \in [0,2]} \mathcal{L}_{b_v}$

\end{algorithmic}
\end{algorithm}

\subsection{Speeding Up Existing BayesOpt Methods}
Below we provide details for incorporating Vecchia GPs into existing BayesOpt frameworks to improve their computational speed.
\subsubsection{Vecchia with TuRBO}\label{sec:vecchiaTurbo}
Trust Region Bayesian optimization (TuRBO) \citep{eriksson2019scalable} is a Bayesian optimization procedure that uses hyper-rectangles called \textit{trust regions}. In TuRBO, independent GPs are fit to data within the trust regions. Query points are chosen using Thompson sampling, and the regions grow or shrink based on how often they contain new optima. TuRBO allows for a single trust region, TuRBO-1 or $t$ simultaneous trust regions, TuRBO-t. TuRBO-1 is based on a single global GP fit to all data, whereas TuRBO-t fits $t$ local GPs, each using only the data within one of the trust regions. The individual trust regions help by allowing greater flexibility in modeling the local behavior of the function and facilitating a more thorough exploration of the entire input space \citep{Eriksson2019}. 

Veccchia fits in the TuRBO framework without much change. Using either TuRBO-1 or TuRBO-t, the same (global) Vecchia GP is shared between all trust regions, and predictions within each trust region are made using the $m$ NNs among all data (i.e., the NNs are not restricted to be in the trust region for either TuRBO-1 or TuRBO-t). By conditioning on the NNs, we emphasize the local behavior of the target function, and by training on all the data we ensure our hyperparameters are well calibrated. When using TuRBO-1, the variance correction of Vecchia amounts to inflating or deflating the predictive variance for points in and around the trust region. For the remainder of this paper, we are referring to TuRBO-1 when we say TuRBO.

\subsubsection{Vecchia With SAASBO}\label{sec:vecchiaSaasbo}

The Sparse AxisAligned Subspace BayesOpt (SAASBO) \citep{eriksson2021high} places horseshoe priors on the lengthscale parameters of the GP surrogate. Using a standard exact GP, the time complexity to generate a sample from the posterior is $\mathcal{O}(dn^3)$. Vecchia GPs can improve on this by again reducing the complexity required to invert the training data covariance matrix, as we make the same NN approximation. We can also reduce the cost to evaluate the likelihood further by using the data-splitting idea from \citet{neal2011mcmc}. That is we can compute the likelihood on batches of the data when we perform MCMC sampling. Since BayesOpt queries points that can be close to one another, we expect this procedure to work well, as this is the setting when \citet{neal2011mcmc} suggests data splitting ought to work well.

\subsubsection{Vecchia With Parallel Acquisition Functions}\label{sec:vecchiaqEI}
For parallel acquisition functions such as q-EI or q-UCB, the query points that maximize the acquisition function can be high-dimensional. Gradients of these acquisition functions can ease this inner optimization problem, but the closed-form expression may not be known. Fortunately, Monte Carlo estimates of the gradient can be used \citep{WilsonMaximizingOptimization}. This approach allows us to estimate the gradient of the utility function with respect to $\textbf{X}$ (the collection of query points) as:
\begin{equation}
    \label{eq:ellGrad}
    \nabla \ell(\textbf{y}) = \frac{d\ell (\textbf{y})}{d\textbf{y}}  \frac{d\textbf{y}}{d\mathcal{M}(\textbf{X)}}\frac{d\mathcal{M}(\textbf{X})}{d\textbf{X}},
\end{equation}
where $\mathcal{M}$ is our surrogate model. Under certain conditions, a Monte Carlo approximation of \eqref{eq:ellGrad} will be unbiased and will converge almost surely to a set of stationary points of q-EI \citep{wang2016parallel}. In short, the following can be used to estimate the gradient of q-EI in a stochastic optimization algorithm and we should expect good performance:
\begin{equation}
    \label{eq:mcGrad}
    \nabla \mathcal{L}(\textbf{X}) \approx \frac{1}{m}\sum_{k=1}^m \nabla \ell(\textbf{y}^k), 
\end{equation}
where $y^k$ are draws from the posterior over $f$. We can then use \eqref{eq:mcGrad} to optimize parallel acquisition functions such as q-EI. Sufficient conditions for the convergence of the stochastic optimization algorithm given in \citet{wang2016parallel} using the approximate gradient given by \eqref{eq:mcGrad} requires, among other things, that the mean and covariance functions of the GP be continuously differentiable for each point in the domain. While this is not guaranteed to be satisfied for Vecchia GPs at a set of points of measure zero, as the data grows the effect of this boundary difference should diminish. In practice we can optimize the acquisition function using standard gradient descent even for small $n$. Additionally, since Vecchia prediction for joint samples is so fast Vecchia GPs can greatly improve the speed of this inner optimization especially for high dimensional problems. 

\section{Numerical Comparisons \label{sec:numerical}}
We evaluated Vecchia GPs with TuRBO on a set of benchmarks. The synthetic functions were Ackley-5, Hartmann-6 and Levy. For the Levy function we used what we referred to as the Levy 55-20 function, where we warp the space with a sigmoid type function and add irrelevent dimensions. Finally, we used two reinforcement-learning examples found in \citet{eriksson2019scalable}: Robot Pushing \citep{wang2018batched} and a Lunar Lander problem in OpenAI's gym environment. 

\subsection{GP Surrogates}
We compared the performance of the following GP surrogates within TuRBO-1:
\begin{description}
    \item[Exact GP:] A standard Gaussian process, which is very slow for large $n$.
    \item[LR-First m:] Low-rank approximation where conditioning set taken as the first $m$ points in the maximin ordering. The conditioning set is the same for all $n$ data points.
    \item[SVI-GP:] The SVI GP \citep{Hensman2013GaussianData} places a Gaussian variational distribution on the outputs of ``pseudo-points.'' The variational parameters, including the location of the $m$ pseudo points are learned with mini-batch gradient descent.  
    \item[Vecchia:] Our method as described in the previous sections, where the likelihood is approximated by conditioning each observation on the $m$ NN subject to the constraint that the neighbors appear before the data point in the maximin ordering.
\end{description}

For the approximate methods, we let $m$ increase polylogarithmically with $n$, as $m = C_0 * \log^2_{10}(n)$. The value of $C_0$ was set to 7.2;  for details about this selection see Appendix \ref{appendix:choosem}.

\subsection{Comparison Measures}

For target functions whose optimum is not known, we simply used the best value found as our measure of performance. For functions whose optimum is known, we used \textit{regret} as our measure of performance. Regret after $t$ function evaluations is given by, $r_t = \min_{1\leq i \leq t}\{y_i - f^*\}$, where $f^*$ is the true minimum and the min is taken over all previously observed function values. Our regret curves are averaged over independent repeats.

\subsection{Results}
\subsubsection{Hartmann-6 and Ackley-5}
\begin{figure}
    \centering
    \includegraphics[width =1\linewidth]{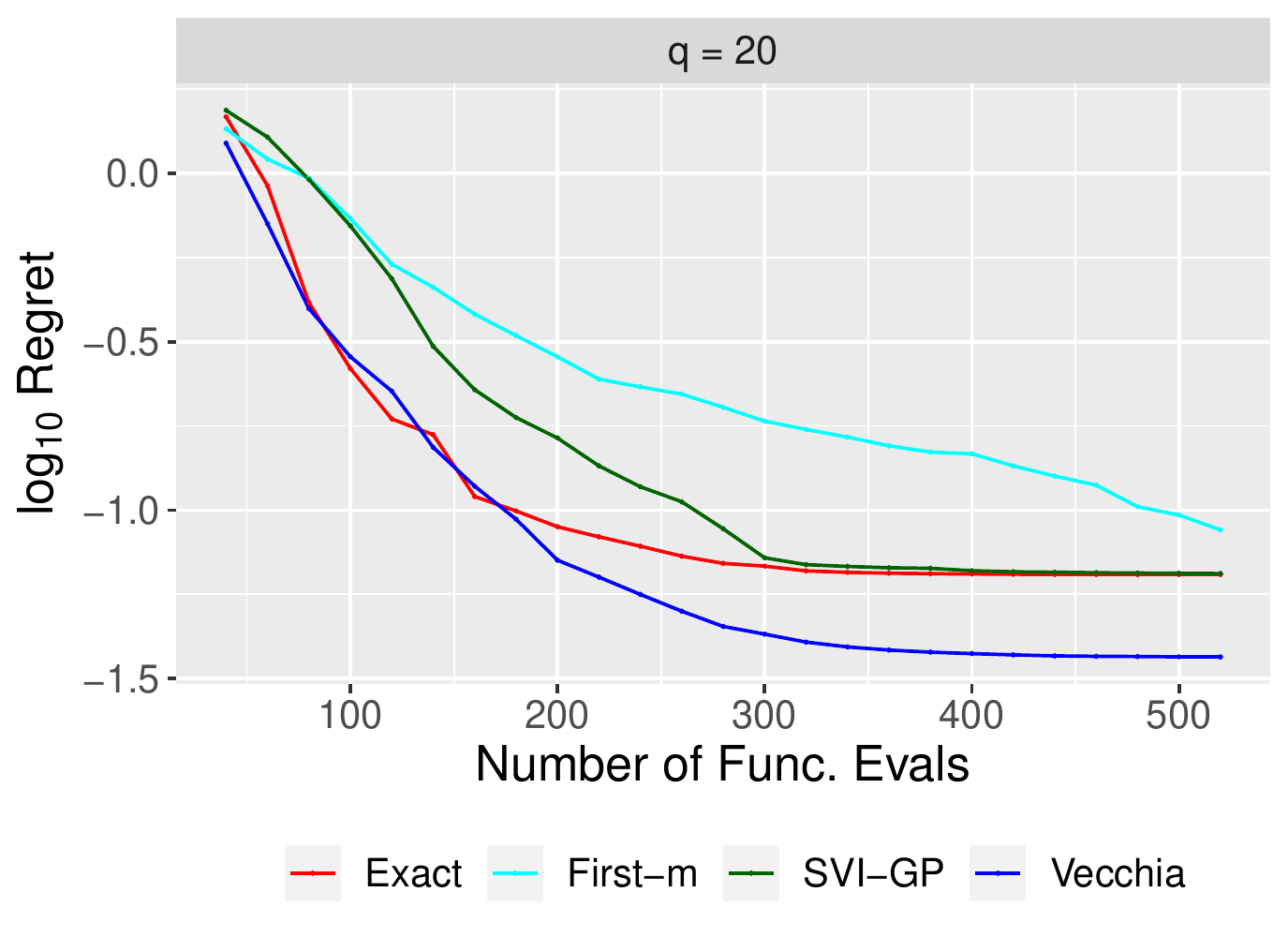}
    \caption{Log regret for Hartmann-6 with batch size $q=20$. Curves are averages of 30 repeats.}
    \label{fig:hartmann6}
\end{figure}

Figure \ref{fig:hartmann6} shows the results for Hartmann-6 with batch size $q = 20$. Vecchia with variance inflation performed better than any other method, including the exact GP. As this function is relatively smooth, in this case Vecchia benefits from the ability to provide joint predictions and from variance inflation. 

\begin{figure}[ht]
    \centering
    \includegraphics[width =1\linewidth]{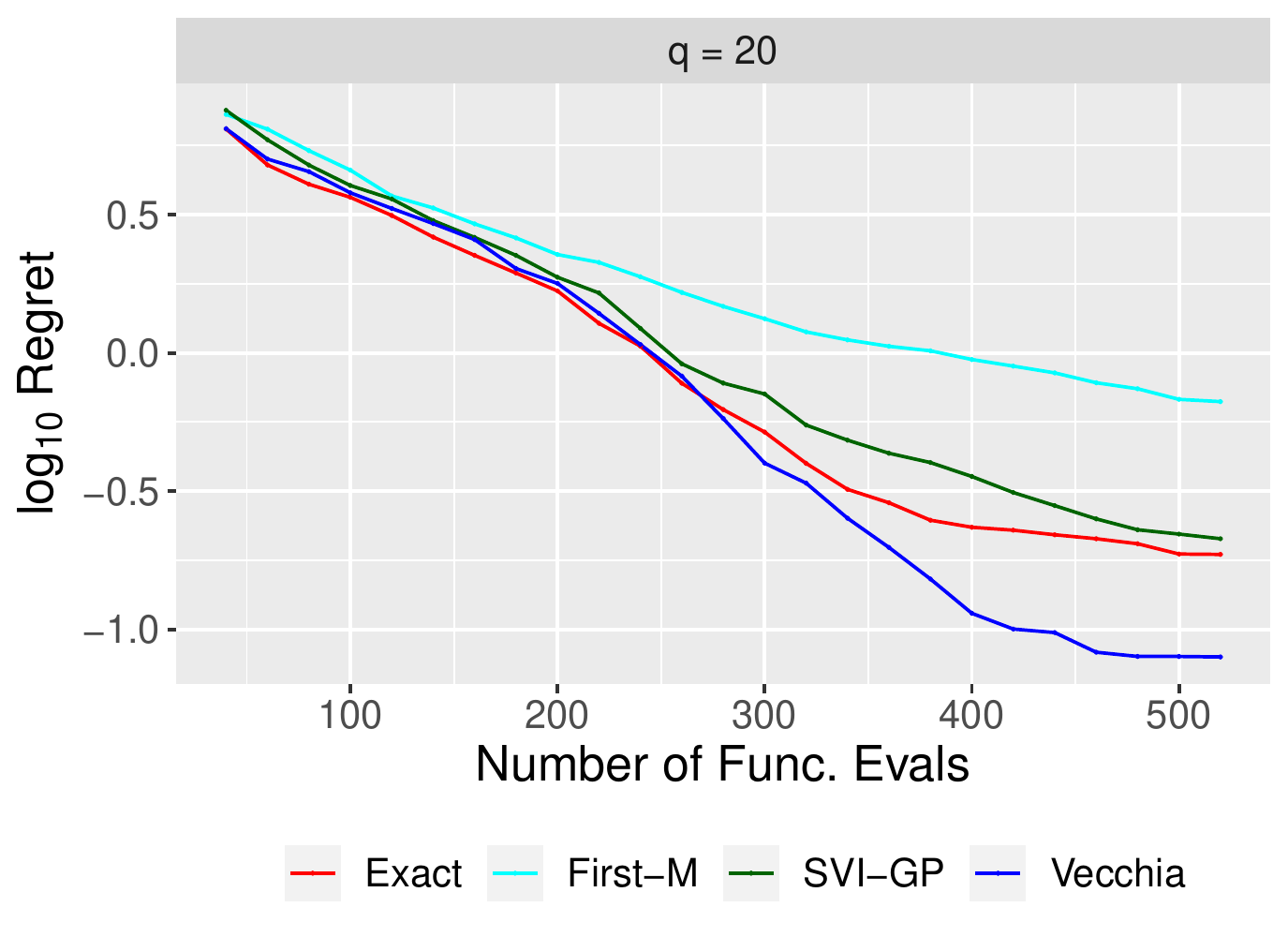}
    \caption{Log regret for Ackley-5 with batch size $q=20$. Curves are averages of 30 repeats.}
    \label{fig:ackley5}
\end{figure}

Figure \ref{fig:ackley5} shows average log regret for Ackley-5 with a batch size of $q = 20$. For Vecchia, we did not use the variance correction. The results exceeded the exact GP by 500 function evaluations and outperformed the other approximate surrogates with ease. 

Both these examples show the flexibility of the Vecchia GP and how modeling choices can be made to improve performance on particular tasks. 

\subsubsection{Warped and embedded Levy}
\begin{figure}[ht]
    \centering
    \includegraphics[width =1\linewidth]{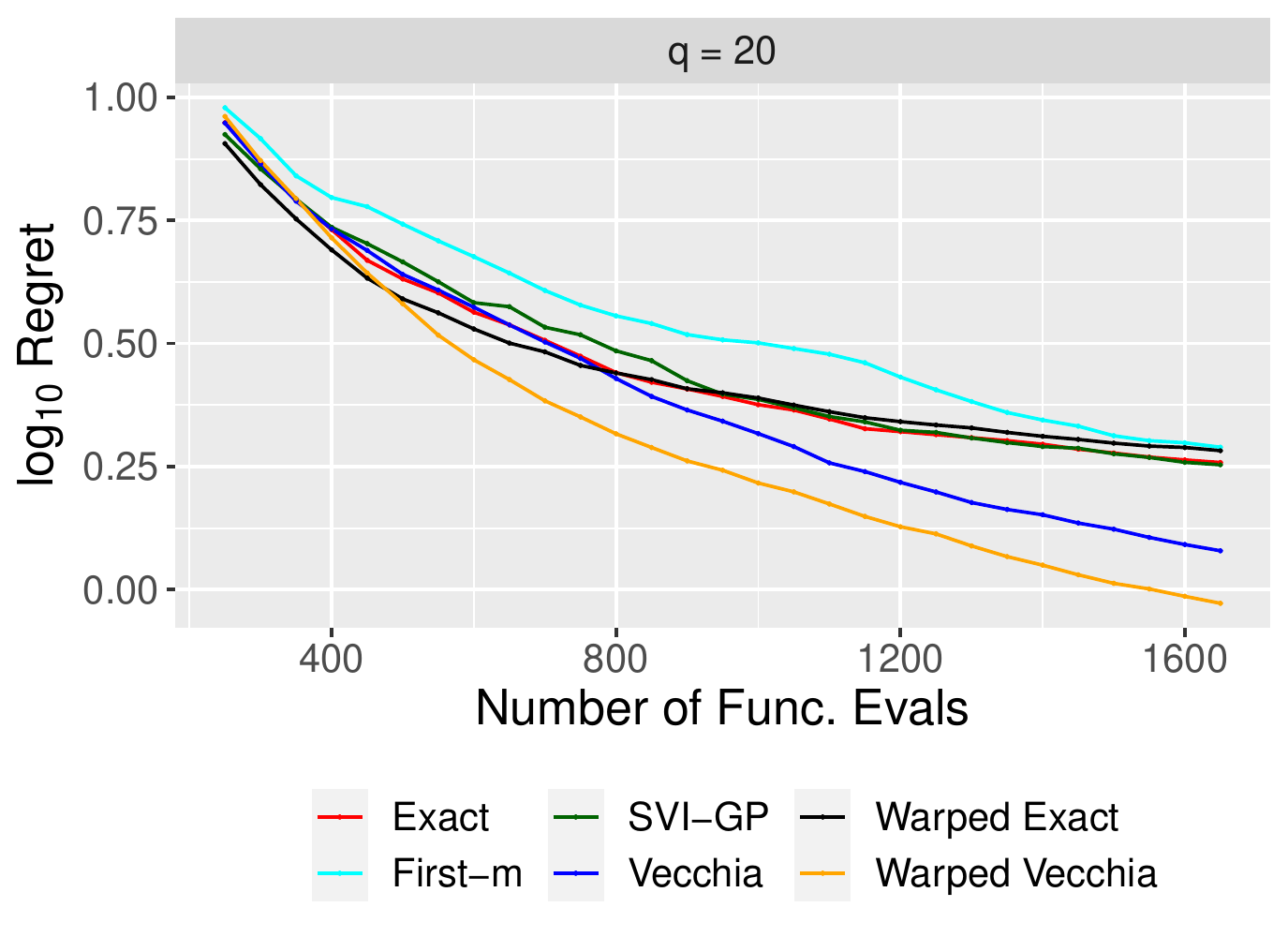}  
    \caption{Log regret for the warped and embedded 20-55 Levy function with batch size $q=50$. Curves were averaged over 30 repeats.}
    \label{fig:levy}
\end{figure}

From Figure \ref{fig:levy}, we see that Vecchia (without warping or variance inflation) did better than all other surrogates; by including the warping capability, the performance of Vecchia improved further. We believe that the warping did not improve the exact GP the same way it improved Vecchia because Vecchia with warping is making the best local approximations of the function possible. Due to the NN conditioning, Vecchia was able to focus on local behavior of this relatively complex function and ignored irrelevant distant parts of the function. 

\subsubsection{Robot Pushing}

In the $14$-dimensional robot-pushing problem \citep{wang2018batched}, there are two robot arms pushing objects towards a target location. We must minimize the distance to the target location. 

\begin{figure}[ht]
    \centering
    \includegraphics[width =1\linewidth]{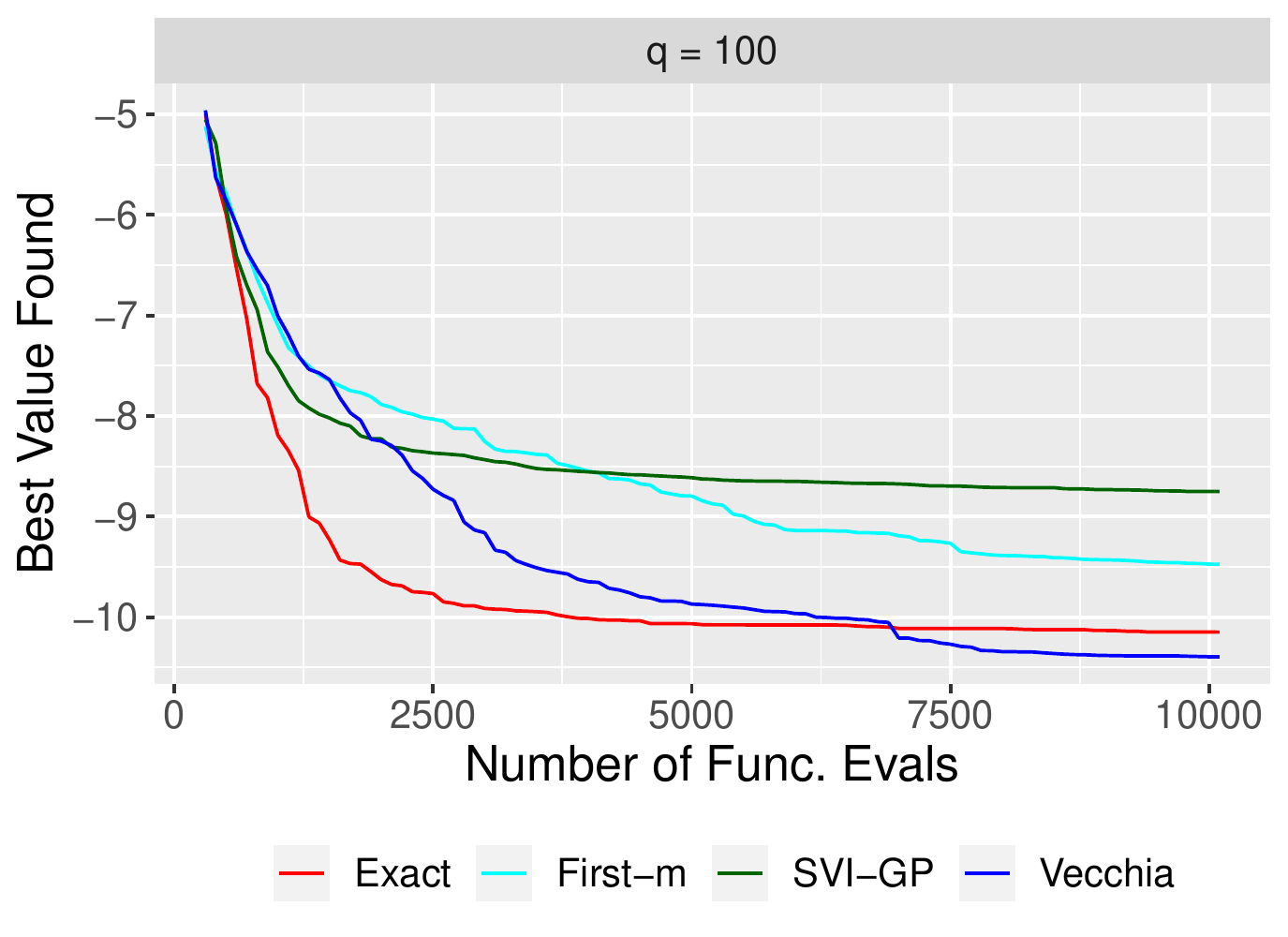}
    \caption{Best value found for 14D Robot Pushing problem with batch size $q=50$. Curves were averaged over 20 repeats.}
    \label{fig:robotPushing}
\end{figure}

As shown in Figure \ref{fig:robotPushing}, Vecchia performed the best among all methods, eventually even outperforming the exact GP for a large number of function evaluations. In this setting, the response was noisy and the target function is highly variable, which is a situation in which using variance inflation for Vecchia GPs within BO works best. 

\subsubsection{Lunar Lander}
We must tune a PID controller \cite{eriksson2019scalable} to successfully land a Lunar module at a fixed point on a set of 50 maps. The results are noisy, because each run of the function is an average over 50 experiments. 

\begin{figure}[ht]
    \centering
    \includegraphics[width =1\linewidth]{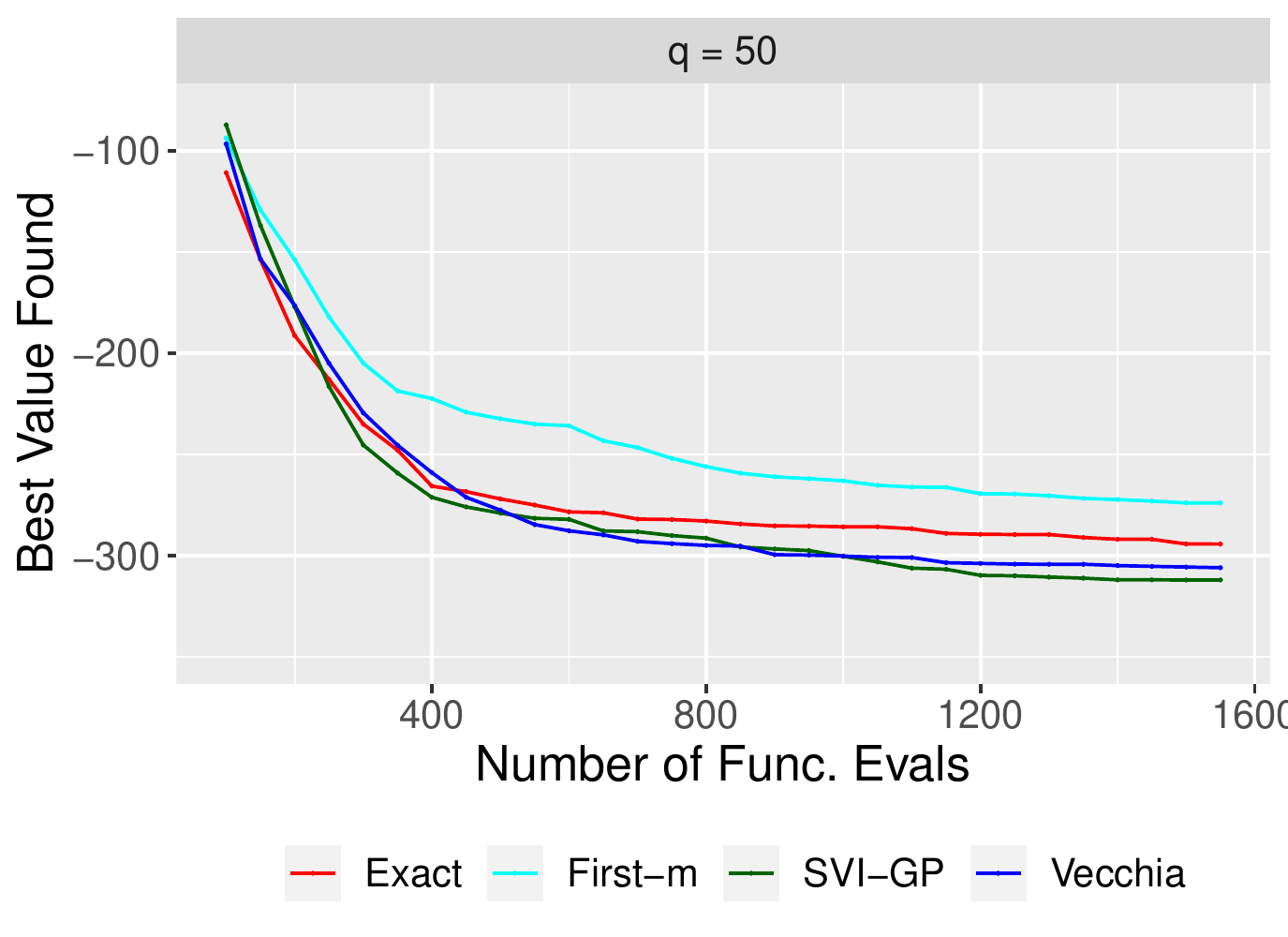}
    \caption{Best value found for 12$D$ Lunar Lander problem with batch size $q=50$. Curves are averaged over 20 repeats.}
    \label{fig:lunar}
\end{figure}

Figure \ref{fig:lunar} shows that scaled Vecchia and SVI-GP performed best, both outperforming the exact GP. Just as with the 14D robot pushing problem, the variance inflation was helpful because the function is noisy. 

\section{Discussion \label{sec:discussion}}

We explored the use of approximate Vecchia GPs for speeding up Bayesian optimization (BO). We described how training of Vecchia GPs can be made efficient through mini-batch gradient descent, approximate nearest neighbors and approximate variable orderings. Additionally, we incorporated Vecchia GPs within the TuRBO BO procedure, and commented on how to use Vecchia GPs for other BO frameworks such as SAASBO and parallel acquisition functions. 

In our numerical experiments on a suite of test functions, we found Vecchia GPs to often be as or even more accurate than other GP surrogates, including the exact GP. This is remarkable, in that, in addition, our Vecchia approximations drastically reduce the computational complexity relative to the exact GP, with a computational cost that largely depends (linearly) on the mini-batch size. As a result, Vecchia can be even computationally cheaper than approximate pseudo-point methods such as SVI-GP, which must optimize over a large number of hyperparameters related to the pseudo-points, especially in high-dimensional input spaces, while Vecchia does not reduce any additional hyperparameters beyond those already present in the exact-GP model to be approximated. We conjecture that Vecchia's accuracy improvements are related to the variance correction and the global-local nature of Vecchia GPs; the nearest neighbor portion of Vecchia focuses on the local behavior of the target function during prediction while using the entire dataset to estimate the hyperparameters from the global information. Together, Vecchia GPs naturally balances local and global information when estimating the target function. 

Our results suggest that Vecchia GPs can be useful for speeding up BO when many evaluations of the target function are feasible and necessary, and we are hopeful that Vecchia approximations can be a standard tool for BO in such settings.

There are several potential avenues for future work. Vecchia GPs are competitive without GPU computing and thus enable fast BO even without access to sophisticated computational resources; however, Vecchia GPs are in principle well suited for GPU-type computations and can thus be sped up further.
For highly noisy target functions, we believe that further accuracy gains can be made by latent Vecchia approximations with incomplete Cholesky decomposition \citep{Schafer2020} within Thompson sampling. Further, correlation-based Vecchia GPs \citep{Kang2021} may be well suited for neural architecture search \citep{kandasamy2018neural}.

\bibliography{mendeley, ref}

\begin{thebibliography}{32}
\providecommand{\natexlab}[1]{#1}
\providecommand{\url}[1]{\texttt{#1}}
\expandafter\ifx\csname urlstyle\endcsname\relax
  \providecommand{\doi}[1]{doi: #1}\else
  \providecommand{\doi}{doi: \begingroup \urlstyle{rm}\Url}\fi

\bibitem[Balandat et~al.(2020)Balandat, Karrer, Jiang, Daulton, Letham, Wilson,
  and Bakshy]{balandat2020botorch}
Balandat, M., Karrer, B., Jiang, D.~R., Daulton, S., Letham, B., Wilson, A.~G.,
  and Bakshy, E.
\newblock {BoTorch: A Framework for Efficient Monte-Carlo Bayesian
  Optimization}.
\newblock In \emph{Advances in Neural Information Processing Systems 33}, 2020.
\newblock URL \url{http://arxiv.org/abs/1910.06403}.

\bibitem[Calandra et~al.(2016)Calandra, Seyfarth, Peters, and
  Deisenroth]{Calandra2016}
Calandra, R., Seyfarth, A., Peters, J., and Deisenroth, M.~P.
\newblock {Bayesian optimization for learning gaits under uncertainty: An
  experimental comparison on a dynamic bipedal walker}.
\newblock \emph{Annals of Mathematics and Artificial Intelligence}, 76\penalty0
  (1-2):\penalty0 5--23, 2 2016.
\newblock ISSN 15737470.
\newblock \doi{10.1007/s10472-015-9463-9}.
\newblock URL
  \url{https://link.springer.com/article/10.1007/s10472-015-9463-9}.

\bibitem[Chapelle \& Li(2011)Chapelle and Li]{Chapelle2011}
Chapelle, O. and Li, L.
\newblock {An Empirical Evaluation of Thompson Sampling}.
\newblock \emph{Advances in Neural Information Processing Systems}, 24, 2011.

\bibitem[Eriksson \& Jankowiak(2021)Eriksson and Jankowiak]{eriksson2021high}
Eriksson, D. and Jankowiak, M.
\newblock {High-Dimensional Bayesian Optimization with Sparse Axis-Aligned
  Subspaces}.
\newblock \emph{arXiv preprint arXiv:2103.00349}, 2021.

\bibitem[Eriksson et~al.(2019{\natexlab{a}})Eriksson, Pearce, Gardner, Turner,
  and Poloczek]{eriksson2019scalable}
Eriksson, D., Pearce, M., Gardner, J., Turner, R.~D., and Poloczek, M.
\newblock {Scalable Global Optimization via Local Bayesian Optimization}.
\newblock In \emph{Advances in Neural Information Processing Systems}, pp.\
  5496--5507, 2019{\natexlab{a}}.

\bibitem[Eriksson et~al.(2019{\natexlab{b}})Eriksson, Pearce, Gardner, Turner,
  and Poloczek]{Eriksson2019}
Eriksson, D., Pearce, M., Gardner, J.~R., Turner, R., and Poloczek, M.
\newblock {Scalable global optimization via local Bayesian optimization}.
\newblock \emph{Neural Information Processing Systems}, 2019{\natexlab{b}}.
\newblock ISSN 23318422.

\bibitem[Frazier(2018)]{Frazier2018AOptimization}
Frazier, P.~I.
\newblock {A Tutorial on Bayesian Optimization}.
\newblock 7 2018.
\newblock URL \url{http://arxiv.org/abs/1807.02811}.

\bibitem[G{\'{o}}mez-Bombarelli et~al.(2018)G{\'{o}}mez-Bombarelli, Wei,
  Duvenaud, Hern{\'{a}}ndez-Lobato, S{\'{a}}nchez-Lengeling, Sheberla,
  Aguilera-Iparraguirre, Hirzel, Adams, and Aspuru-Guzik]{gomez2018automatic}
G{\'{o}}mez-Bombarelli, R., Wei, J.~N., Duvenaud, D., Hern{\'{a}}ndez-Lobato,
  J.~M., S{\'{a}}nchez-Lengeling, B., Sheberla, D., Aguilera-Iparraguirre, J.,
  Hirzel, T.~D., Adams, R.~P., and Aspuru-Guzik, A.
\newblock {Automatic chemical design using a data-driven continuous
  representation of molecules}.
\newblock \emph{ACS central science}, 4\penalty0 (2):\penalty0 268--276, 2018.

\bibitem[Guinness(2018)]{Guinness2016a}
Guinness, J.
\newblock {Permutation and grouping methods for sharpening Gaussian process
  approximations}.
\newblock \emph{Technometrics}, 60\penalty0 (4):\penalty0 415--429, 2018.
\newblock \doi{10.1080/00401706.2018.1437476}.
\newblock URL \url{http://arxiv.org/abs/1609.05372}.

\bibitem[Hensman et~al.(2013)Hensman, Fusi, and
  Lawrence]{Hensman2013GaussianData}
Hensman, J., Fusi, N., and Lawrence, N.~D.
\newblock {Gaussian Processes for Big Data}.
\newblock 9 2013.
\newblock URL \url{http://arxiv.org/abs/1309.6835}.

\bibitem[Jegou et~al.(2010)Jegou, Douze, and Schmid]{jegou2010product}
Jegou, H., Douze, M., and Schmid, C.
\newblock {Product quantization for nearest neighbor search}.
\newblock \emph{IEEE transactions on pattern analysis and machine
  intelligence}, 33\penalty0 (1):\penalty0 117--128, 2010.

\bibitem[Johnson et~al.(2017)Johnson, Douze, and J{\'{e}}gou]{JDH17}
Johnson, J., Douze, M., and J{\'{e}}gou, H.
\newblock {Billion-scale similarity search with GPUs}.
\newblock \emph{arXiv preprint arXiv:1702.08734}, 2017.

\bibitem[Kandasamy et~al.(2015)Kandasamy, Schneider, and
  P{\'{o}}czos]{kandasamy2015high}
Kandasamy, K., Schneider, J., and P{\'{o}}czos, B.
\newblock {High dimensional Bayesian optimisation and bandits via additive
  models}.
\newblock In \emph{International conference on machine learning}, pp.\
  295--304. PMLR, 2015.

\bibitem[Kandasamy et~al.(2018)Kandasamy, Neiswanger, Schneider, Poczos, and
  Xing]{kandasamy2018neural}
Kandasamy, K., Neiswanger, W., Schneider, J., Poczos, B., and Xing, E.
\newblock {Neural architecture search with bayesian optimisation and optimal
  transport}.
\newblock \emph{arXiv preprint arXiv:1802.07191}, 2018.

\bibitem[Kang \& Katzfuss(2021)Kang and Katzfuss]{Kang2021}
Kang, M. and Katzfuss, M.
\newblock {Correlation-based sparse inverse Cholesky factorization for fast
  Gaussian-process inference}.
\newblock \emph{arXiv:2112.14591}, 2021.

\bibitem[Katzfuss \& Guinness(2021)Katzfuss and Guinness]{Katzfuss2017a}
Katzfuss, M. and Guinness, J.
\newblock {A general framework for Vecchia approximations of Gaussian
  processes}.
\newblock \emph{Statistical Science}, 36\penalty0 (1):\penalty0 124--141, 2021.
\newblock \doi{10.1214/19-STS755}.
\newblock URL \url{http://arxiv.org/abs/1708.06302}.

\bibitem[Katzfuss et~al.(2020)Katzfuss, Guinness, Gong, and
  Zilber]{Katzfuss2018}
Katzfuss, M., Guinness, J., Gong, W., and Zilber, D.
\newblock {Vecchia approximations of Gaussian-process predictions}.
\newblock \emph{Journal of Agricultural, Biological, and Environmental
  Statistics}, 25\penalty0 (3):\penalty0 383--414, 2020.
\newblock \doi{10.1007/s13253-020-00401-7}.

\bibitem[Katzfuss et~al.(2022)Katzfuss, Guinness, and Lawrence]{Katzfuss2020}
Katzfuss, M., Guinness, J., and Lawrence, E.
\newblock {Scaled Vecchia approximation for fast computer-model emulation}.
\newblock \emph{SIAM/ASA Journal on Uncertainty Quantification}, accepted,
  2022.
\newblock URL \url{http://arxiv.org/abs/2005.00386}.

\bibitem[Letham et~al.(2019)Letham, Karrer, Ottoni, and Bakshy]{Letham2019}
Letham, B., Karrer, B., Ottoni, G., and Bakshy, E.
\newblock {Constrained Bayesian optimization with noisy experiments}.
\newblock \emph{Bayesian Analysis}, 14\penalty0 (2):\penalty0 495--519, 2019.
\newblock \doi{10.1214/18-BA1110}.
\newblock URL \url{http://arxiv.org/abs/1706.07094}.

\bibitem[Maddox et~al.(2021)Maddox, Stanton, and
  Wilson]{maddox2021conditioning}
Maddox, W.~J., Stanton, S., and Wilson, A.~G.
\newblock {Conditioning Sparse Variational Gaussian Processes for Online
  Decision-making}.
\newblock \emph{arXiv preprint arXiv:2110.15172}, 2021.

\bibitem[McIntire et~al.(2016)McIntire, Ratner, and Ermon]{mcintire2016sparse}
McIntire, M., Ratner, D., and Ermon, S.
\newblock {Sparse Gaussian Processes for Bayesian Optimization.}
\newblock In \emph{UAI}, 2016.

\bibitem[Mockus(1989)]{Mockus1989}
Mockus, J.
\newblock \emph{{Bayesian Approach to Global Optimization}}.
\newblock Kluwer, Dordrecht, NL, 1989.
\newblock ISBN 9789401068987.

\bibitem[Molga \& Smutnicki(2005)Molga and Smutnicki]{molga2005test}
Molga, M. and Smutnicki, C.
\newblock {Test functions for optimization needs}.
\newblock \emph{Test functions for optimization needs}, 101:\penalty0 48, 2005.

\bibitem[Neal \& {others}(2011)Neal and {others}]{neal2011mcmc}
Neal, R.~M. and {others}.
\newblock {MCMC using Hamiltonian dynamics}.
\newblock \emph{Handbook of markov chain monte carlo}, 2\penalty0
  (11):\penalty0 2, 2011.

\bibitem[Sch{\"{a}}fer et~al.(2021)Sch{\"{a}}fer, Katzfuss, and
  Owhadi]{Schafer2020}
Sch{\"{a}}fer, F., Katzfuss, M., and Owhadi, H.
\newblock {Sparse Cholesky factorization by Kullback-Leibler minimization}.
\newblock \emph{SIAM Journal on Scientific Computing}, 43\penalty0
  (3):\penalty0 A2019--A2046, 2021.
\newblock \doi{10.1137/20M1336254}.

\bibitem[Stein(2014)]{Stein2013a}
Stein, M.~L.
\newblock {Limitations on low rank approximations for covariance matrices of
  spatial data}.
\newblock \emph{Spatial Statistics}, 8:\penalty0 1--19, 5 2014.
\newblock ISSN 22116753.
\newblock \doi{10.1016/j.spasta.2013.06.003}.
\newblock URL
  \url{http://linkinghub.elsevier.com/retrieve/pii/S2211675313000390
  https://linkinghub.elsevier.com/retrieve/pii/S2211675313000390}.

\bibitem[Swersky et~al.(2013)Swersky, Snoek, and Adams]{Swersky2013}
Swersky, K., Snoek, J., and Adams, R.~P.
\newblock {Multi-Task Bayesian Optimization}.
\newblock Technical report, 2013.

\bibitem[Thompson(1933)]{Thompson1933}
Thompson, W.~R.
\newblock {On the likelihood that one unknown probability exceeds another in
  the view of evidence of two samples}.
\newblock \emph{Biometrika}, 25\penalty0 (3-4):\penalty0 285--294, 12 1933.
\newblock ISSN 0006-3444.
\newblock \doi{10.1093/BIOMET/25.3-4.285}.
\newblock URL \url{https://academic.oup.com/biomet/article/25/3-4/285/200862}.

\bibitem[Vecchia(1988)]{Vecchia1988}
Vecchia, A.
\newblock {Estimation and model identification for continuous spatial
  processes}.
\newblock \emph{Journal of the Royal Statistical Society, Series B},
  50\penalty0 (2):\penalty0 297--312, 1988.
\newblock URL \url{http://www.jstor.org/stable/10.2307/2345768}.

\bibitem[Wang et~al.(2016)Wang, Clark, Liu, and Frazier]{wang2016parallel}
Wang, J., Clark, S.~C., Liu, E., and Frazier, P.~I.
\newblock {Parallel bayesian global optimization of expensive functions}.
\newblock \emph{arXiv preprint arXiv:1602.05149}, 2016.

\bibitem[Wang et~al.(2018)Wang, Gehring, Kohli, and Jegelka]{wang2018batched}
Wang, Z., Gehring, C., Kohli, P., and Jegelka, S.
\newblock {Batched large-scale Bayesian optimization in high-dimensional
  spaces}.
\newblock In \emph{International Conference on Artificial Intelligence and
  Statistics}, pp.\  745--754. PMLR, 2018.

\bibitem[Wilson et~al.(2018)Wilson, Hutter, and
  Deisenroth]{WilsonMaximizingOptimization}
Wilson, J.~T., Hutter, F., and Deisenroth, M.~P.
\newblock {Maximizing acquisition functions for Bayesian optimization}.
\newblock Technical report, 2018.

\end{thebibliography}
\bibliographystyle{icml2022}

%%%%%%%%%%%%%%%%%%%%%%%%%%%%%%%%%%%%%%%%%%%%%%%%%%%%%%%%%%%%%%%%%%%%%%%%%%%%%%%
%%%%%%%%%%%%%%%%%%%%%%%%%%%%%%%%%%%%%%%%%%%%%%%%%%%%%%%%%%%%%%%%%%%%%%%%%%%%%%%
% APPENDIX
%%%%%%%%%%%%%%%%%%%%%%%%%%%%%%%%%%%%%%%%%%%%%%%%%%%%%%%%%%%%%%%%%%%%%%%%%%%%%%%
%%%%%%%%%%%%%%%%%%%%%%%%%%%%%%%%%%%%%%%%%%%%%%%%%%%%%%%%%%%%%%%%%%%%%%%%%%%%%%%
\newpage
\appendix
\onecolumn

\section{Appendix}
\subsection{Approximate Variable Orderings}\label{appendix:approxMaxMin}
The function below gives an approximate maxmin ordering which works well empirically, and can be run in parallel. We assume the inputs have been processed to the correct space (e.g. warped, scaled by inverse lengthscales). Additionally, we assume there is a function that can perform the exact maxmin ordering, such as Algorithm C.1 in the supplementary material of \cite{Schafer2020}.

\begin{algorithm}
\caption{Approximate MaxMin Ordering}\label{alg:approxMaxMin}
\begin{algorithmic}[1]
\STATE {\bfseries Input:} data $\textbf{x}$, parameters $MaxSubsetSize$
\STATE $N \gets length(\textbf{x})$
  \IF{$N > MaxSubsetSize$}
    \STATE $order1 \gets  ApproximateMaxMin(\textbf{x}[0:floor(N/2)], MaxSubsetSize)$ \COMMENT{Run in Parallel.}
    \STATE $order2 \gets  ApproximateMaxMin(\textbf{x}[floor(N/2):N], MaxSubsetSize)$ \COMMENT{Run in Parallel.}
    \STATE $order = concatenate(order1, order2)$
  \ELSE
    \STATE $order = GetExactMaxMin(\bm x)$
  \ENDIF
  
\STATE Return $order$
\end{algorithmic}
\end{algorithm}

\subsection{Prediction}
Following \citet{Katzfuss2018} we consider the response first ordering (RF) using RF-standard (RF-stand). In RF-stand, the $j$th prediction location conditions on any previously ordered data point, including another prediction location. This means we take into account the posterior covariance between some prediction locations if they are closer together than to training data. For clarity we use the subscribt $r$ to refer to the observed data and the subcript $p$ to refer to predictions. 

For prediction we form two matrices $\bm U_{r,p}$ and $\bm U_{p,p}$ whose entries can be computed using equation \ref{eqn:uappendix}. Using the $\bm U$ matrices we can get the posterior mean and the Cholesky factor of the posterior variance with complexity dependent on the number of prediction locations, and not the sample size of the training data. See Section 4.1 in \citet{Katzfuss2018} for details. 

The Cholesky factor of the posterior precision for prediction locations is given by:
\begin{equation}
    \bm L_p= (\bm U_{p,p}^{-1})^T
\end{equation}
The posterior predictive mean is given by 
\begin{equation}
    \bm \mu_p = (\bm U_{p,p}^T)^{-1}\bm U_{r, p}^T \bm y_{1:n}.
\end{equation}

\subsubsection{U Matrix}\label{appendix:UMatrix}
We before consider the matrix $\textbf{U}$, whose elements are given by:
\begin{equation}\label{eqn:uappendix}
\textbf{U}_{j,i}=
    \begin{cases}
        d_i^{-1/2}, & i=j\\
        -b_i^{(j)}d_i^{-1/2}, & j\in g(i)\\
        0, & \text{otherwise}, 
    \end{cases}
\end{equation}

When we allow the prediction locations to condition on other prediction locations then there will be non-zero elements in the matrix $\textbf{U}_{pp}$, and there will be columns with less than $m$ non-zero elements in the matrix $\textbf{U}_{n,p}$. The significance of this becomes obvious when we consider the posterior mean, since a location may condition on less than $m$ observations. The prediction will instead rely on the value that was predicted at other unknown locations. 

This works well when the number of observations is much greater than the number of predictions. If we restrict ourselves to condition only on observed data then we ignore the joint nature of the posterior, but gain speed as all the operations can be done in parallel.

\subsubsection{Kernel and Warping}\label{sec:warpedKernel}
When the target function cannot be modeled by a stationary GP, it can help to first warp the input space through a non-linear invertible function.
%\citep{snelson2004warped, snoek2014input}. 
To incorporate this tool into the Vecchia framework, we consider a general kernel that allows for warping,
\begin{equation}
    K(\bm x,\bm x') = \tilde K(\|w(\bm x)-w(\bm x')\|),
\end{equation}x
where $\|\cdot\|$ denotes Euclidean distance and $\tilde K$ is an isotropic covariance function. We assume the warping function to operate element-wise on each of the $d$ input dimensions, such that $w(\bm x) = \big(w_1(\bm x_1),\ldots,w_d(\bm x_d))^\top$, where $w_j(\bm x_j) = \omega_{\bm \phi_j}(\bm x_j)/\lambda_j$; here, $\lambda_j$ is the range parameter for the $j$th input dimension, and $\omega_{\bm \phi}$ is bijective, continuous, and differentiable with respect to $\bm \phi$. For our numerical experiments, we used the Kumarswamy CDF \citep{balandat2020botorch}, $\omega_{a,b}(\bm x) = 1-(1-\bm x^a)^b$; when its parameters are set to $a=b=1$, \eqref{eq:warpingkernel} becomes the well-known automatic relevance determination (ARD) kernel.

For approximating a GP with a warping kernel, we propose the warped Vecchia approximation. Specifically, in $\eqref{eq:vecchia}$, we carry out the maximin ordering of $\bm y_{1:n}$ and the selection of NN conditioning sets $c(i)$ based on Euclidean distance between the corresponding warped inputs, $w(\bm x_1),\ldots,w(\bm x_n)$. For prediction of a new variable $y^*_i$ at $\bm x_i^*$, we also select $c^*(i)$ as the $m$ NNs to $w(\bm x_i^*)$ among $w(\bm x_1),\ldots,w(\bm x_n)$. In the special case of linear warping with $\omega(\bm x) = \bm x$, this warped Vecchia approach is equivalent to the scaled Vecchia approximation for emulating computer experiments \citep{Katzfuss2020}.

\subsection{Experiment Details}
Below are extra details on the experimental setup.
\subsubsection{Experimental Setup}\label{sec:experimentSetup}
During every round, each GP surrogate model under consideration was fit using all of the available data. Thompson sampling was used within each TuRBO trust region, with the input locations chosen from a Sobol sequence. We evaluated the GP at $\min(5000, \max(5000, 100 d))$ locations $q$-separate times. We chose the batch of $q$ elements to be the locations with the $q$ highest simulated values within Thompson sampling. For all surrogates, we used a Mat\'ern2.5 covariance function for $\tilde K$ in \eqref{eq:warpingkernel}, assuming linear warping (i.e., $a=b=1$) for all examples except the Levy function. For approximate surrogate models, we chose $m$ as described in Section \ref{appendix:choosem}. The input space for all test functions and problems was mapped to $[0,1]^d$. All results were averaged over independent replicates, with starting points again chosen using a Sobol sequence. 

\subsubsection{Choosing \texorpdfstring{$m$}{m}}\label{appendix:choosem}

For the approximate surrogates, we grow $m$ polylogarithmically with $n$, $m = C_0 * \log^2_{10}(n)$. To simplify our comparison between surrogate models, we chose a single value of $C_0$ for all our experiments. To choose this $m$, we looked at the regret vs number of samples when we used the Vecchia approximation with EI as our acquisition function and compared with the regret curve using the exact GP. When using the Six-Hump camel function we found a $C_0$ value of around 7.2 is the lowest for which Vecchia tracked the exact solution closely. We excluded the Six-Hump camel function from the comparison of regret curves to put the competing approximations on more even footing. 

\subsubsection{Levy 55-20 function}\label{appendix:levy5520}
We used a modified version of the Levy-20 function to show the benefit of warping with Vecchia. We call this function Levy 55-20, and it is simply the Levy-20 function with 35 extra irrelevant dimensions. The hyper cube on $[0,1]^{55}$ was warped using $w(\bm x)=S(4 * \bm x -1)$ ($S(\cdot)$ is the sigmoid function) before being scaled to $[-5,5]^{55}$. Only the first 20 dimensions are used to evaluate the Levy-20 function.

\end{document}